\documentclass[10pt,twocolumn,letterpaper]{article}

\usepackage{cvpr}              

\usepackage{graphicx}
\usepackage{amsmath}
\usepackage{amssymb}
\usepackage{booktabs}
\usepackage[accsupp]{axessibility}  

\usepackage[pagebackref,breaklinks,colorlinks]{hyperref}

\usepackage[capitalize]{cleveref}
\crefname{section}{Sec.}{Secs.}
\Crefname{section}{Section}{Sections}
\Crefname{table}{Table}{Tables}
\crefname{table}{Tab.}{Tabs.}

\usepackage{booktabs}       
\usepackage{amsfonts}       
\usepackage{nicefrac}       
\usepackage{microtype}      

\usepackage{amsthm}
\usepackage{amsmath}
\usepackage{amssymb}
\usepackage{dsfont}
\usepackage{graphicx}
\usepackage{mathtools}
\usepackage{subcaption}
\usepackage{bbm}
\usepackage{color}
\usepackage{xspace}

\usepackage{graphicx}
\usepackage{booktabs}

\usepackage{algorithm}
\usepackage{algpseudocode}
\DeclareMathOperator*{\argmin}{arg\,min} 
\usepackage[pagebackref,breaklinks,colorlinks]{hyperref}

\newcommand{\RR}{\mathbb{R}}

\newcommand{\defeq}{\vcentcolon=}

\def\cvprPaperID{5623} 
\def\confName{CVPR}
\def\confYear{2022}

\begin{document}

\title{Exploiting Temporal Relations on Radar Perception for Autonomous Driving}

\author{Peizhao Li$^1$\thanks{Work done during the internship at MERL}$\ $, Pu Wang$^2$, Karl Berntorp$^2$, Hongfu Liu$^1$\\
$^1$Brandeis University, $^2$Mitsubishi Electric Research Laboratories\\
{\tt\small \{peizhaoli,hongfuliu\}@brandeis.edu, \{pwang,berntorp\}@merl.com}
}
\maketitle

\begin{abstract}
We consider the object recognition problem in autonomous driving using automotive radar sensors. Comparing to Lidar sensors, radar is cost-effective and robust in all-weather conditions for perception in autonomous driving. However, radar signals suffer from low angular resolution and precision in recognizing surrounding objects. To enhance the capacity of automotive radar, in this work, we exploit the temporal information from successive ego-centric bird-eye-view radar image frames for radar object recognition. We leverage the consistency of an object's existence and attributes (size, orientation, \etc.), and propose a temporal relational layer to explicitly model the relations between objects within successive radar images. In both object detection and multiple object tracking, we show the superiority of our method compared to several baseline approaches.
\end{abstract}

\vspace{-5mm}
\section{Introduction}
\label{sec:intro}

Autonomous driving utilizes sensing technology for robust dynamic object perception, and sequentially uses the perception for reliable and safe vehicle decision-making~\cite{yurtsever2020survey}. Among various perception sensors, camera and Lidar are the two dominant ones exploited for surrounding object recognition. The camera provides semantically rich visual features of traffic scenarios, while Lidar provides high-resolution point clouds that can capture the reflection from objects. Compared with camera and Lidar, radar enjoys the following unique advantages when applied in automotive applications. Primarily operating at $77$ GHz, radar transmits electromagnetic waves at a millimeter wavelength to estimate the range, velocity, and angle of objects. At such a wavelength, it can penetrate or diffract around tiny particles in conditions such as rain, fog, snow, and dust, and offer long-range perception in these adverse weather conditions~\cite{ZengNickolaou14}. In contrast, laser sent by Lidar at a much shorter wavelength may bounce off these tiny particles, which leads to a significantly reduced operating range. Compared with the camera, radar is also resilient to light conditions, \eg, night and sun glare. Furthermore, radar offers a cost-effective and reliable option to complement other sensors. For the cost of Lidar, according to an aggressive estimate by Luminar, is expected to be the range of \$500 - \$1000~\cite{lidar_price}. In contrast, automotive radar is expected to be less than \$100 in 2022~\cite{radar_price}. However, as a disadvantage of radar-assisted automotive perception, a high angular resolution in the azimuth and elevation domains are indispensable. In recent open-access automotive radar datasets, an azimuth resolution of $1^{\circ}$ becomes available, while the elevation resolution is still lagging behind. With $1^{\circ}$ azimuth resolution, semantic features for objects in a short range, \eg, corners and shapes, can be observed, while an object at far distances can still be blurred due to the cross-range resolution. In summary, the capability of localizing and identifying objects for radar is still falling behind from full-level autonomous driving.

\begin{figure}[t]
    \centering
    \includegraphics[width=\columnwidth]{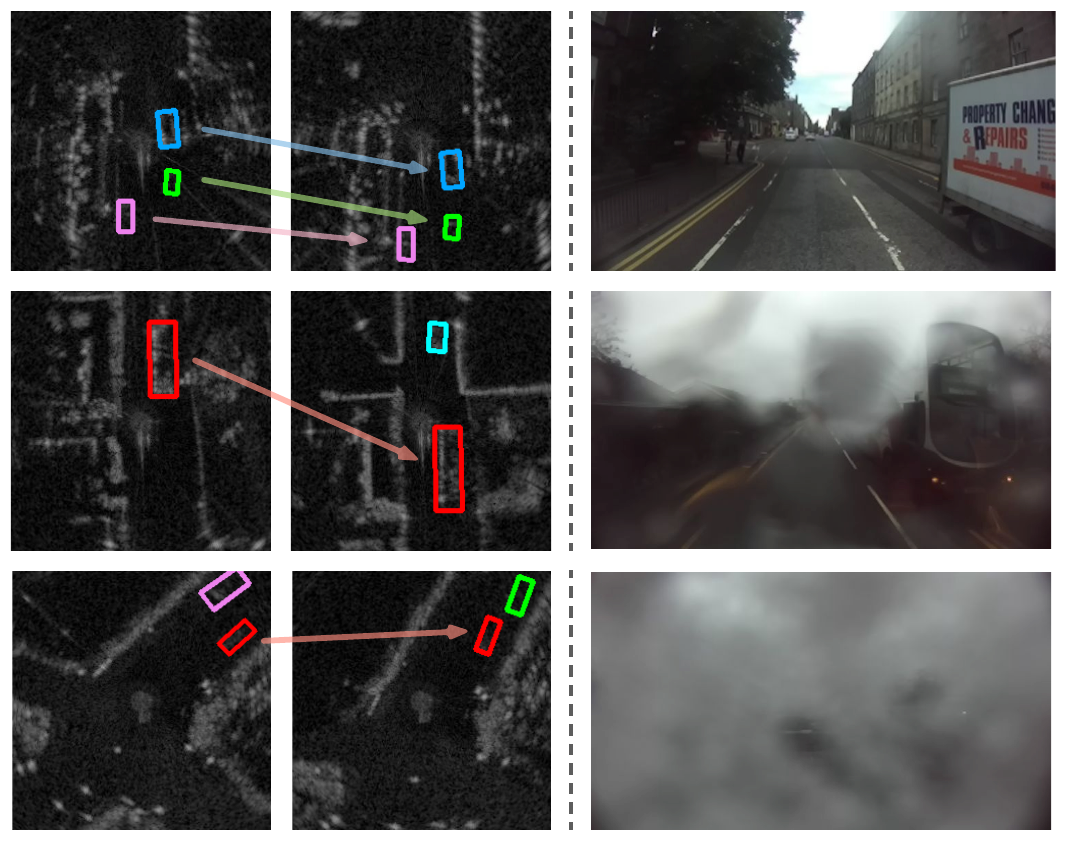}\vspace{-2mm}
    \caption{Showcasing of two successive radar images and the corresponding camera recording from \textit{Radiate} dataset~\cite{sheeny2020radiate}. From top to bottom, we display examples in the normal, foggy, and snowy weather. The bounding boxes are the ground-truth annotations of objects where its color implies the object ID. The plotted arrows show the consistency of the object's appearance and attributes within a short time period, \eg, length, width, and orientation.}
    \label{fig:tem}
    \vspace{-5.5mm}
\end{figure}

Some recent efforts have been taken to leverage and enhance automotive radar for object recognition from an algorithmic perspective. \cite{major2019vehicle} proposes a deep-learning approach using range-azimuth-doppler measurement. \cite{qian2021robust} detects objects via synchronous radar and Lidar signals. Similarly, \cite{yang2020radarnet,lim2019radar} exploit the multi-modal sensing fusion. Besides deep learning, Bayesian learning has also attempted to solve extended object tracking with radar point clouds~\cite{yao2021extended,xia2021learning}. The above works mainly focus on multi-modal sensing fusion for robust perception~\cite{qian2021robust,yang2020radarnet,lim2019radar}. Differently, in this paper, we take our attempt to enhance the perception only using radar information, which requires fewer perception resources and avoids a complicated synchronized process for signals among multi-modal sensors.


In this paper, we consider ego-centric bird-eye-view radar point clouds presented in a Cartesian frame, where pixel values indicate the strength of reflections. We develop an approach to enhance radar perception using temporal information. Based on the observation in Fig.~\ref{fig:tem}, we assume that the same objects detected by radar within successive frames are consistent and share almost the same attributes, such as the object's existence, length, orientation, \etc. As a result, the detection at one frame can be facilitated by a previous/future frame through object-level correlations. To compensate for the blurriness and low angular resolution raised by radar sensors, we involve temporality and incorporate customized temporal relational layers to explicitly handle the object-level relations across successive frames. The temporal relational layer takes feature vectors at the potential object's centers and conducts a temporal as well as a self-attention over the object features which are wrapped with their locality. Colloquially, this layer links temporally similar objects and transmits their representations, and is akin to feature smoothing. Hence, temporal relational layers could insert the inductive bias from object temporal consistency. Afterward, the object heatmap (indicating the center of objects) and relevant attributes are inferred upon the updated feature representation from temporal relational layers.

In this work, we consider the object recognition problem using radar in autonomous driving, which is a crucial alternative sensing technology that owes unique advantages. We underline major contributions of our work as follows:
\vspace{-2mm}
\begin{itemize}
    \item We facilitate the radar perception with additional temporal information to compensate for the blurriness and low angular resolution raised by radar sensors.
    \vspace{-2mm}
    \item We design a customized temporal relational layer, where the networks are inserted with an inductive bias that the same object in successive frames should share consistent appearance and attributes.
    \vspace{-2mm}
    \item We evaluate our method in object detection and multiple object tracking on \textit{Radiate} dataset. With the comprehensive comparison to baseline methods, we show the consistent improvements brought by our method.
\end{itemize}


\vspace{-3mm}
\section{Radar Perception: Background}

\begin{figure}[t]
   \begin{center}
   \begin{tabular}{cc}
   \includegraphics[width=1.\linewidth]{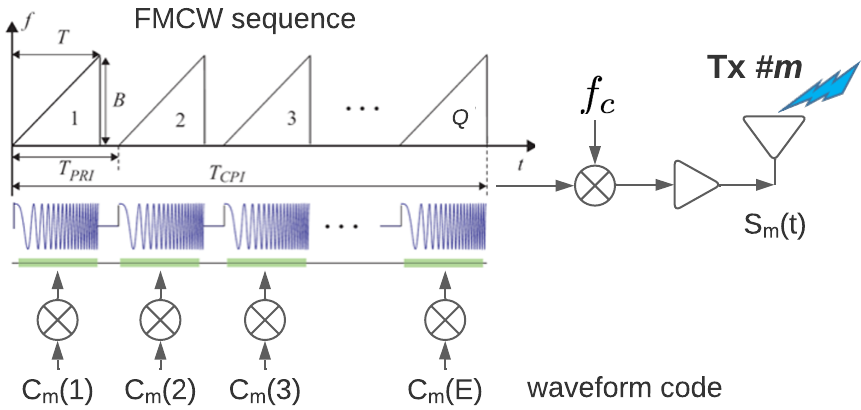} \\
   (a) Transmitter (Tx) \\
   \includegraphics[width=1.\linewidth]{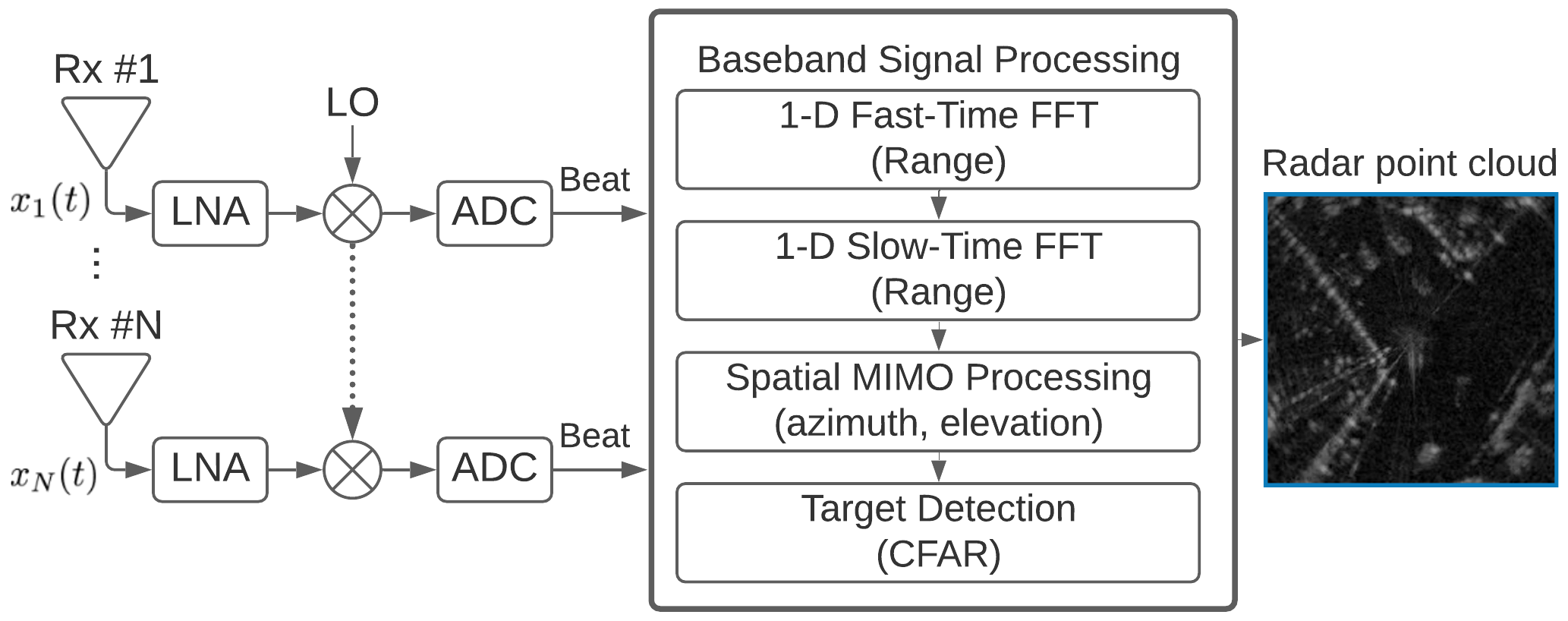} \\
   (b) Receiver (Rx)
   \end{tabular}
   \caption{FMCW-based automotive radar.} 
   \label{fig:FMCW}
   \end{center}
   \vspace{-8mm}
\end{figure}

Automotive radar dominantly uses frequency modulated continuous waveform (FMCW) to detect objects and generate point clouds over multiple physical domains. As shown in Fig.~\ref{fig:FMCW} (a), it transmits a sequence of FMCW pulses through one of its $M$ transmitting antennas:
\begin{align}\label{st}
s_m(t) = \sum\limits_{q=0}^{Q-1} c_m(q) s_p\left(t-nT_{\text{PRI}} \right) e^{j 2\pi f_c t},
\end{align}
where $m$ and $q$ are the indices for transmitting antenna and pulse, $T_{\text{PRI}}$ is pulse repetition interval, $f_c$ is the carrier frequency (e.g., $79$ GHz), and $s_p(t)$ is baseband FMCW waveform (shown as the sinusoids in Fig.~\ref{fig:FMCW} (a)).

An object at a range of $R_0$ with a radial velocity $v_t$ and a far-field spatial angle (\ie azimuth, elevation, or both) induces amplitude attenuation and phase modulation to the received FMCW signal at each of $N$ receiver RF chains (including the low noise amplifier (LNA), local oscillator (LO), and analog-to-digital converter (ADC)) of Fig.~\ref{fig:FMCW} (b). The induced modulation from the target is captured by the baseband signal processing block (including fast Fourier transforms (FFTs) over range, Doppler, and spatial domains) in Fig.~\ref{fig:FMCW} (b). All these processes lead to a multi-dimensional spectrum. With the constant false alarm rate (CFAR) detection step that compares the spectrum with an adaptive threshold, radar point clouds are generated in the range, Doppler, azimuth, and elevation domains \cite{LiStoica08,WangBoufounos20,BilikLongman19}.


Considering the computing and cost constraints, automotive radar manufactures may define the radar point clouds in a subset of the full four dimensions. For instance, traditional automotive radar generates detection points in the range-Doppler domain, whereas some produce the points in the range-Doppler-azimuth plane~\cite{RamasubramanianGinsburg17}. In \textit{Radiate} dataset~\cite{sheeny2020radiate} considered in this paper, the radar point cloud is defined in the range-azimuth plane with a $360^\circ$ field view. The resulting polar-coordinate point cloud is further transformed into an ego-centric Cartesian coordinate system, then a standard voxelization can convert the point cloud into an image.


\begin{figure*}[t]
    \centering
    \includegraphics[width=2\columnwidth]{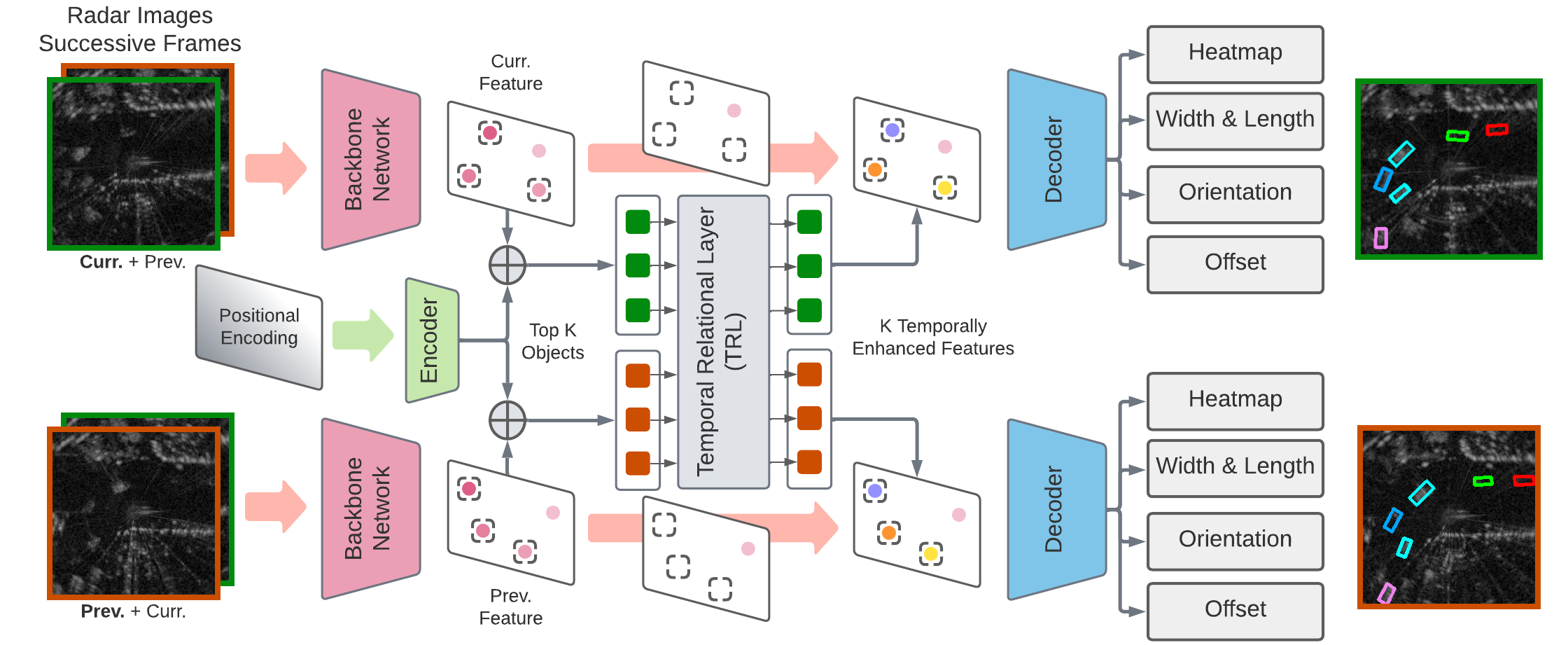}
    \vspace{-4mm}
    \caption{The framework of radar object recognition with temporality. Viewing from left to right, our method takes two consecutive radar frames and extracts the temporal feature from each frame. Then, we select features that could be potential objects and learn the temporal consistency between them. Finally, several regression objectives are conducted upon the updated features for training.}
    \label{fig:framework}
    \vspace{-6mm}
\end{figure*}

\section{Radar Perception with Temporality}

We present our framework in Fig.~\ref{fig:framework}. Corresponding to Fig.~\ref{fig:framework} from top to bottom, in the subsequent sections, we introduce the temporal feature extraction from two successive frames, the temporal relational layers, the learning method, followed by the extension to multiple object tracking.


\vspace{-4.5mm}
\paragraph{Notation} We clarify the following notations. $\theta$ denotes the learnable parameters in neural networks, and for simplification, we unify the notations of parameters with $\theta$ for all modules. We use a bracket following a three-dimensional matrix to represent the feature gathering process at certain coordinates. Consider a feature representation $Z\in\RR^{C\times H\times W}$ with $C$, $H$, and $W$ represent channel, height, and width, respectively. Let $P$ represent a coordinate $(x,y)$ or a set of two-dimensional coordinates $\{(x,y)\}_K$ with cardinality equal to $K$ and $x, y\in\RR$. $Z[P]$ means taking the feature at a coordinate system indicated by $P$ along width and height dimensions, with the returned features in $\RR^C$ or $\RR^{K\times C}$.


\subsection{Temporal Feature Extraction}

Denote a single radar frame as $I\in\RR^{1\times H\times W}$. We concatenate two successive radar images: a current frame and its previous frame, along the channel dimension to involve temporal information at the input level. The channel-concatenated temporal input image for the current and previous frames can be respectively written as $I_{c+p}$ and $I_{p+c}\in\RR^{2\times H\times W}$. The order of `current' $c$ and `previous' $p$ in the subscript indicates the feature-concatenating order of these two frames. We obtain the feature representations for the two frames by forwarding the formulated inputs through a backbone neural network $\mathcal{F}_\theta(\cdot)$:
\begin{equation}
    Z_c \defeq \mathcal{F}_\theta(I_{c+p}),\ \ Z_p \defeq \mathcal{F}_\theta(I_{p+c}).
\end{equation}
The backbone network $\mathcal{F}_\theta(\cdot)$ is built in standard deep convolutional neural networks (\eg, ResNet), and model parameters are shared for processing two inputs $I_{p+c}$ and $I_{c+p}$.

To jointly involve high-level semantics and low-level finer details in feature representations, we build skip connections between features at different scales in neural networks. Specifically, for one skip connection, we up-sample the pooled feature from a deep layer to align its size with the feature from previous shallow layers via bilinear interpolation. A list of operations including convolution, non-linear activation, and batch normalization are afterward applied to the up-sampled feature. Next, the up-sampled features are concatenated with those from shallow layers along the channel dimension. Three skip connections are inserted into the networks to drive the features embrace semantics at four different levels. The final feature representation from the backbone neural networks are resulted in $Z_{c}, Z_{p}\in\RR^{C\times \frac{H}{s}\times \frac{W}{s}}$, where $s$ is the down-sampling ratio over the spatial dimension. We add an illustrative figure in Appendix A.


\subsection{Modeling Object Temporal Relations}

We design a temporal relational layer to model the correlation and consistency between potential objects in successive frames. The temporal relational layer receives multiple feature vectors from the two frames with each vector representing a potential object in a radar image. We apply a filtering module $\mathcal{G}_\theta^{\text{pre-hm}}: \RR^{C\times \frac{H}{s}\times \frac{W}{s}} \rightarrow \RR^{1\times \frac{H}{s}\times \frac{W}{s}}$ on features $Z_c$ and $Z_p$ to select top $K$ potential object features for the relational modeling. The set of coordinates $P_c$ for potential objects in $Z_c$ is obtained via the following equation:
\begin{equation}\label{eq:pos}
    P_c \defeq \{ (x, y)\mid\mathcal{G}_\theta^{\text{pre-hm}}(Z_{c})_{xy} \geq [\mathcal{G}_\theta^{\text{pre-hm}}(Z_{c})]_K\},
\end{equation}
where $[\mathcal{G}_\theta^{\text{pre-hm}}(Z_{c})]_K$ is the $K$-th largest value in $\mathcal{G}_\theta^{\text{pre-hm}}(Z_{c})$ over the spatial space $\frac{H}{s}\times \frac{W}{s}$, and the subscript $xy$ denotes taking value at coordinate $(x,y)$. Clearly, the cardinality of $P_c$ is $|P_c| = K$. By substituting $Z_p$ into Eq.~\eqref{eq:pos}, $P_p$ for $Z_p$ can be obtained similarly. We do not include features from all coordinates into the temporal relational layer due to that the computational complexity of the subsequent attention mechanism grows quadratically towards the value $K$.

By taking the coordinate sets $P_c$ and $P_p$ into feature representations, we have the selective feature matrix as:
\begin{equation}
    \textbf{H}_c \defeq Z_c[P_c],\ \textbf{H}_p \defeq Z_p[P_p].
\end{equation}
Sequentially, let $\textbf{H}_{c+p} \defeq \begin{bmatrix} \textbf{H}_{c},\textbf{H}_p\\ \end{bmatrix}^\top\in\RR^{2K\times C}$ denote the matrix concatenation of top-$K$ selected features in the two frames that forms the input to the temporal relational layer.

We supplement the positional encoding into feature vectors before passing $\textbf{H}_{c+p}$ into the temporal relational layer. The reason is that Convolutional neural networks do not encompass absolute positional information into output feature representation since CNNs enjoy the translational invariance property. However, the position is crucial in object temporal relations because objects at a certain spatial distance in two successive frames are more likely to be associated and would share similar object's attributes. The spatial distance between the same object is conditional on the frame rate and vehicle's motion, and can be learned through a data-driven approach. Denote $\textbf{H}_{c+p}^{\text{pos}}\in\RR^{2K\times (C+D_\text{pos})}$ as the feature supplemented by the positional encoding via feature concatenation, where $D_\text{pos}$ is the dimension of positional encoding. Positional encoding is projected from the normalized 2D coordinate $(x, y)$ that takes values in $[0,1]$ via linear mappings.

Having the formulations above, we have our main operation for modeling the relations across frames. For a single $l$-th temporal relational layer, we use a superscript $l$ to denote the input feature and $l+1$ to denote the output feature:
\begin{equation}\label{eq:att}
    \textbf{H}_{c+p}^{l+1} = \text{softmax}\left(\frac{\textbf{M} + q(\textbf{H}_{c+p}^{l, \text{pos}}) k(\textbf{H}_{c+p}^{l, \text{pos}})^\top}{\sqrt{d}}\right)v(\textbf{H}_{c+p}^{l}),
\end{equation}
where $q(\cdot)$, $k(\cdot)$, and $v(\cdot)$ are linear transformation layers applied to features and are referred as, respectively, query, keys, and values. $d$ is the dimension of query and keys and is used to scale the dot product between them. The masking matrix $\textbf{M}\in\RR^{2K\times 2K}$ is defined as:
\begin{equation}
    \textbf{M} \defeq \sigma\cdot \left(\begin{bmatrix} \textbf{1}_{K,K}, \textbf{0}_{K,K}\\ \textbf{0}_{K,K}, \textbf{1}_{K,K}\\\end{bmatrix} - \mathbbm{1}_{2K} \right),
\end{equation}
where $\textbf{1}_{K,K}$ is the all-one matrix with size $K\times K$, $\textbf{0}_{K,K}$ is the all-zero matrix with size $K\times K$, $\mathbbm{1}_{2K}$ is the identity matrix of size $2K$, and $\sigma$ is a negative constant which is set to $-(1\mathrm{e}{+10})$ in our implementation to guarantee a near-zero value in the output through softmax. The diagonal matrices of $\textbf{1}_{K,K}$ disable the attention between features from the same frame, while the off-diagonal matrices of $\textbf{0}_{K,K}$ allow the cross-frame attention. Also, the identity matrix $\mathbbm{1}_{2K}$ unlocks the object self-attention. The logic behind self-attention is that the same object co-occurrence cannot always be guaranteed in successive frames since an object can move out of the scope, thereby self-attention is desirable when an object is missing in only one frame. Noticeably, the positional encoding is only attached to keys and query but not to values, so the output feature does not involve locality. Other technical details follows the design of Transformer~\cite{vaswani2017attention}, and here we omit the detailed descriptions for simplification.

After executing the object temporal attention across frames in Eq.~\eqref{eq:att}, we sequentially apply a feed-forward function that consists of two linear layers, layer normalization, and shortcut on features. The relational modeling is built with multiple temporal relational layers with the identical design. At the end, we split the updated features $\textbf{H}_{c}^{l+1}$ and $\textbf{H}_{p}^{l+1}$ from $\textbf{H}_{c+p}^{l+1}$ and refill the feature vector to $Z_c$ and $Z_p$ in the corresponding spatial coordinates from $P_c$ and $P_p$. Regressions in the next subsection are conducted on top of the refilled feature representations.

\vspace{-5mm}
\paragraph{Discussion} The above feature operations share some similarities with Transformer~\cite{vaswani2017attention}. Transformer is designed for language representation learning, intending to map the words into a similar latent representation if two words are sharing correlations among the training corpus, including the co-existence, word positions, and semantics. The multi-head attention operations in the stacked architecture can be understood as smoothing over the feature of semantically similar words~\cite{dong2021attention,gong2021improve,Li_2021_CVPR}. In our context, the feature of objects with an identical ID in successive frames should be correlated and share a similar latent representation. This is particularly crucial since the latent representation store all object-relevant attributes and will be used for the subsequent decoding purpose, as elaborated in Section~\ref{sec:learn}. The smoothing over two feature vectors of the same object in successive frames satisfies our basic temporal consistency assumption, and can enhance the detection when the object information is partially lost in one frame due to the blurriness from radar.


\subsection{Learning}\label{sec:learn}

We pick the object's center coordinates from the heatmap, and learn its attributes (\ie the width, length, orientation, and center coordinate offset) from feature representations through regression.

\vspace{-5mm}
\paragraph{Heatmap} To localize objects, the 2D coordinate of a peak value in the heatmap is considered as the center of an object. The heatmap is obtained by a module $\mathcal{G}_\theta^{\text{hm}}: \RR^{C\times \frac{H}{s}\times \frac{W}{s}} \rightarrow \RR^{1\times \frac{H}{s}\times \frac{W}{s}}$ followed by a sigmoid function. We generate the ground-truth heatmap by placing the 2D radial basis function (RBF) kernel on the center of every ground-truth object, while the parameter $\sigma$ in the RBF kernel is set proportional to the object's width and length. Considering the sparsity of objects in radar images, we use focal loss~\cite{lin2017focal} to balance the regression of ground-truth centers and background, and drive the predicted heatmap to approximate the ground-truth heatmap. Let $h_i$ and $\hat{h}_i$ denote the ground-truth and predicted value at $i$-th coordinate, $N$ the total number of values in the heatmap, we express the focal loss as:
\begin{equation}
\begin{split}
    L_h \defeq& -\frac{1}{N}\sum_i \big(\mathds{1}_{h_i = 1}(1 - \hat{h}_i)^\alpha\log(\hat{h}_i) \\
    &+ \mathds{1}_{h_i \neq 1}(1 - h_i)^\beta \hat{h}_i^\alpha \log(1 - \hat{h}_i) \big),
\end{split}
\end{equation}
where $\alpha$ and $\beta$ are hyper-parameters and are chosen empirically with $2$ and $4$, respectively, following the prior work~\cite{yi2021oriented}. The same loss function is conducted for $\mathcal{G}_\theta^{\text{pre-hm}}$ to rectify the feature selection of the relational modeling. During inference, a threshold is set on the heatmap to distinguish the object center from backgrounds. Non-maximum suppression is applied to avoid excessive bounding boxes.

\vspace{-4mm}
\paragraph{Width \& Length} We predict the width and length of an oriented bounding box from the feature vector positioned at the center coordinate in the feature map through another regression head $\mathcal{G}_\theta^{\text{b}}: \RR^C \rightarrow \RR^2$. Let $P_{\text{gt}}^k$ denote the coordinate $(x, y)$ of the center of $k$-th ground-truth object, $b^k$ the ground-truth vector containing width and length of $k$-th object, and $Z$ a unified notation for $Z_c$ and $Z_p$. We have:
\begin{equation}
    L_\text{b} \defeq \frac{1}{N}\sum_{k=1}^{N} \text{Smooth}_{L_1} \left(\lVert\mathcal{G}_\theta^\text{b}(Z[P_{\text{gt}}^k]) - b^k \rVert \right),
\end{equation}
where the $L_1$ smooth loss is defined as:
\begin{equation}
    \text{Smooth}_{L_1}(x) \defeq
    \begin{cases}
      0.5x^2 & \text{if}\ |x|<1\\
      |x|-0.5 & \text{otherwise}.\\
    \end{cases}     
\end{equation}

\vspace{-4mm}
\paragraph{Orientation} All vehicles are presented with an orientation in the bird-eye-view image. An angle range in $[0^{\circ}, 360^{\circ})$ can be measured by the deviation between the object's orientation and the boresight direction of the ego vehicle. We regress the sine and cosine values of the angle $\vartheta$ via $\mathcal{G}_\theta^{\text{r}}: \RR^C \rightarrow \RR^2$:
\begin{equation}
    L_\text{r} \defeq \frac{1}{N}\sum_{k=1}^{N} \text{Smooth}_{L_1}(\lVert \mathcal{G}_\theta^\text{r}(Z[P_{\text{gt}}^k]) - (\sin(\vartheta), \cos(\vartheta)) \rVert).
\end{equation}
During the inference stage, the orientation can be predicted by ${\sin(\hat{\vartheta})}$ and ${\cos(\hat{\vartheta})})$ via $\arctan({\sin(\hat{\vartheta})} / {\cos(\hat{\vartheta})})$.

\vspace{-4mm}
\paragraph{Offset} Down sampling in the backbone networks could incur a center coordinate shift for every object. The center coordinates in the heatmap are integers while the true coordinates are likely to be off the heatmap grids due to the spatial down sampling. To compensate for the shift, we calculate a ground-truth offset for the $k$-th object as:
\begin{equation}
    o^k \defeq \left(\frac{c_x^k}{s} - \left[\frac{c_x^k}{s}\right],\ \frac{c_y^k}{s} - \left[\frac{c_y^k}{s} \right] \right),
\end{equation}
where $c_x^k$ and $c_y^k$ is the $k$-th center coordinate, $s$ is the down sampling ratio, and the bracket $[\cdot]$ is the rounding operation to an integer. Having $\mathcal{G}_\theta^{\text{o}}: \RR^C \rightarrow \RR^2$, the regression for center positional offset can be similarly expressed as:
\begin{equation}
    L_\text{o} \defeq \frac{1}{N}\sum_{k=1}^{N} \text{Smooth}_{L_1}(\lVert \mathcal{G}_\theta^\text{o}(Z[P_{\text{gt}}^k]) - o^k \rVert).
\end{equation}

\vspace{-6mm}
\paragraph{Training} All above regression functions compose the final training objective by a linear combination:
\begin{equation}
    \min_\theta\  L \defeq L_h + L_b + L_r + L_o.
\end{equation}
We omit the balanced factors for each term for simplification.

For each training step, our training procedure calculates the loss $L$ and does the backward for both the current and previous frame simultaneously. Standing at the current frame, objects in the current frame receives information from the past for object recognition. On the other hand, from the previous frame perspective, objects utilize the temporal information from the immediate future frame. Therefore, the optimization can be viewed as a \textit{bi-directional} backward-forward training towards two successive frames. For now, we do not extend the current framework to multiple frames, since an intermediate frame do not have a proper concatenated order of input images for temporal feature extraction (neither from past to future or nor from future to past) and would reduce the training efficiency.


\subsection{Extending to Multiple Object Tracking}

Our framework can be easily extended to online multiple object tracking by adapting a similar tracking procedure as in~\cite{zhou2020tracking}. For multiple object tracking, we add a regression head to the center feature vector to predict a 2D moving offset between the center of an object holding the same tracking ID in current and previous frames. We simply use Euclidean distance to accomplish the association in tracking decoding. We defer a detailed illustration and algorithm for Multiple Object Tracking to Appendix B.


\section{Experiment}


\subsection{Experimental Setup}

\begin{table*}
  \caption{Experimental results of object detection on \textit{Radiate} dataset. TRL is the abbreviation of `temporal relational layer.'}\vspace{-3mm}
  \label{tab:det}
  \centering
  \setlength\tabcolsep{6.5pt}
  {
  \begin{tabular}{lccccccc}
    \toprule
    
    & \multicolumn{3}{c}{Split: train good weather} & & \multicolumn{3}{c}{Split: train good and bad weather} \\
    \cmidrule{2-4}\cmidrule{6-8}
    & mAP@0.3 & mAP@0.5 & mAP@0.7 & & mAP@0.3 & mAP@0.5 & mAP@0.7 \\
    
    \midrule
    
    RetinaNet-OBB-ResNet18 & 52.50\scriptsize{$\pm$ 1.81} & 37.83\scriptsize{$\pm$ 1.82} & 8.46\scriptsize{$\pm$ 0.61} & & 49.44\scriptsize{$\pm$ 1.32} & 31.57\scriptsize{$\pm$ 1.54} & 6.97\scriptsize{$\pm$ 1.24} \\
    RetinaNet-OBB-ResNet34 & 50.79\scriptsize{$\pm$ 3.10} & 35.61\scriptsize{$\pm$ 3.35} & 7.67\scriptsize{$\pm$ 1.71} & & 48.09\scriptsize{$\pm$ 3.85} & 31.10\scriptsize{$\pm$ 3.37} & 6.93\scriptsize{$\pm$ 1.60} \\
    RetinaNet-OBB-ResNet34-T. & 52.52\scriptsize{$\pm$ 4.68} & 37.30\scriptsize{$\pm$ 3.35} & 8.75\scriptsize{$\pm$ 1.50} & & 42.95\scriptsize{$\pm$ 3.46} & 24.50\scriptsize{$\pm$ 3.72} & 3.98\scriptsize{$\pm$ 1.55} \\
    CenterPoint-OBB-EfficientNetB4 & 61.15\scriptsize{$\pm$ 1.23} & 51.43\scriptsize{$\pm$ 1.45} & 20.31\scriptsize{$\pm$ 1.73} & & 54.97\scriptsize{$\pm$ 2.59} & 42.37\scriptsize{$\pm$ 2.14} & 13.15\scriptsize{$\pm$ 0.98} \\
    CenterPoint-OBB-ResNet18 & 58.69\scriptsize{$\pm$ 3.09} & 49.41\scriptsize{$\pm$ 2.94} & 19.02\scriptsize{$\pm$ 1.80} & & 55.83\scriptsize{$\pm$ 3.28} & 44.48\scriptsize{$\pm$ 3.19} & 14.43\scriptsize{$\pm$ 2.56} \\
    CenterPoint-OBB-ResNet34 & 59.42\scriptsize{$\pm$ 1.92} & 50.17\scriptsize{$\pm$ 1.91} & 18.93\scriptsize{$\pm$ 1.46} & & 53.92\scriptsize{$\pm$ 3.44} & 42.81\scriptsize{$\pm$ 3.04} & 13.43\scriptsize{$\pm$ 1.92} \\
    BBAVectors-ResNet18 & 59.38\scriptsize{$\pm$ 3.47} & 50.53\scriptsize{$\pm$ 2.07} & 19.72\scriptsize{$\pm$ 1.10} & & 56.84\scriptsize{$\pm$ 3.45} & 45.43\scriptsize{$\pm$ 2.87} & 15.07\scriptsize{$\pm$ 1.76} \\
    BBAVectors-ResNet34 & 60.88\scriptsize{$\pm$ 1.79} & 51.26\scriptsize{$\pm$ 1.99} & 19.86\scriptsize{$\pm$ 1.36} & & 55.87\scriptsize{$\pm$ 2.90} & 44.61\scriptsize{$\pm$ 2.57} & 14.67\scriptsize{$\pm$ 1.45} \\
    
    \midrule
    
    Ours-EfficientNetB4-w/o TRL & 60.77\scriptsize{$\pm$ 0.97} & 50.93\scriptsize{$\pm$ 1.27} & 20.31\scriptsize{$\pm$ 1.73} & & 54.97\scriptsize{$\pm$ 2.59} & 42.37\scriptsize{$\pm$ 2.14} & 13.15\scriptsize{$\pm$ 0.98} \\
    Ours-EfficientNetB4-w. TRL & 61.59\scriptsize{$\pm$ 1.54} & 50.98\scriptsize{$\pm$ 1.52} & 17.91\scriptsize{$\pm$ 1.48} & & 55.28\scriptsize{$\pm$ 2.32} & 43.05\scriptsize{$\pm$ 2.63} & 13.48\scriptsize{$\pm$ 2.01} \\
    Ours-ResNet18-w/o TRL & 57.48\scriptsize{$\pm$ 4.82} & 47.90\scriptsize{$\pm$ 4.77} & 16.85\scriptsize{$\pm$ 2.98} & & 55.64\scriptsize{$\pm$ 2.32} & 44.48\scriptsize{$\pm$ 2.76} & 15.10\scriptsize{$\pm$ 1.68} \\
    Ours-ResNet18-w. TRL & 62.79\scriptsize{$\pm$ 2.01} & 53.11\scriptsize{$\pm$ 1.96} & 20.57\scriptsize{$\pm$ 1.47} & & \textbf{58.87}\scriptsize{$\pm$ 3.31} & \textbf{46.42}\scriptsize{$\pm$ 3.24} & \textbf{15.59}\scriptsize{$\pm$ 2.31} \\
    Ours-ResNet34-w/o TRL & 60.98\scriptsize{$\pm$ 1.89} & 49.98\scriptsize{$\pm$ 2.28} & 18.89\scriptsize{$\pm$ 1.46} & & 57.21\scriptsize{$\pm$ 3.76} & 45.93\scriptsize{$\pm$ 3.52} & 15.51\scriptsize{$\pm$ 2.71} \\
    Ours-ResNet34-w. TRL & \textbf{63.63}\scriptsize{$\pm$ 2.08} & \textbf{54.00}\scriptsize{$\pm$ 2.16} & \textbf{21.08}\scriptsize{$\pm$ 1.66} & & 56.18\scriptsize{$\pm$ 4.27} & 43.98\scriptsize{$\pm$ 3.75} & 14.35\scriptsize{$\pm$ 2.15} \\
    
    \bottomrule
  \end{tabular}\vspace{-4mm}
  }
\end{table*}

\begin{table}
  \caption{Comparison on object detection to~\cite{sheeny2020radiate}. Results of~\cite{sheeny2020radiate} are directly copied from the original paper.}\vspace{-3mm}
  \label{tab:det_full}
  \centering
  \setlength\tabcolsep{2pt}
  {
  \begin{tabular}{lc}
    \toprule
    split: train good weather & mAP@0.5  \\
    \midrule
    FasterRCNN-ResNet50~\cite{sheeny2020radiate} & 45.31 \\
    FasterRCNN-ResNet101~\cite{sheeny2020radiate} & 45.84 \\
    Ours-ResNet18-w. TRL & 48.02 \\
    Ours-ResNet34-w. TRL & \textbf{48.66} \\
    \bottomrule
  \end{tabular}\vspace{-6mm}
  }
\end{table}

\paragraph{Dataset} We use the radar dataset \textit{Radiate}~\cite{sheeny2020radiate} in our experiments for the following reasons: (1) it contains high-resolution radar images; (2) it provides well-annotated oriented bounding boxes with tracking IDs for objects; and (3) it records various real driving scenarios in adverse weather. \textit{Radiate} is consist of video sequences recorded in adverse weather including sun, night, rain, fog, and snow. The driving scenarios vary from the motorway to the urban. The data format radar images generated from point clouds, where pixel values indicate the strength of radar signal reflections. \textit{Radiate} adopts mechanically scanning Navtech CTS350-X radar, providing $360^{\circ}$ high-resolution range-azimuth images at 4 Hz. Currently, the radar does not afford doppler or velocity information. The whole dataset has in total 61 sequences and we follow the official 3 splits: train in good weather (31 sequences, 22383 frames, only in good weather, sunny or overcast), train good and bad weather (12 sequences, 9749 frames, both good and bad weather conditions), and test (18 sequences, 11305 frames, all kinds of weather conditions). We separately train models on the former two training sets and evaluate on the test set. Numerical results from both two splits are reported. We also comprehensively review other public radar datasets and discuss why currently they are not feasible for our experiments in Section~\ref{sec:relate}.

\vspace{-5mm}
\paragraph{Baseline} We implement several detectors, which have been well demonstrated in visual object detection for comparison. These detectors include: Faster-RCNN~\cite{ren2015faster}, RetinaNet~\cite{lin2017focal}, CenterPoint~\cite{zhou2019objects}, and BBAVectors~\cite{yi2021oriented}. The comparison is conducted with different backbone networks~\cite{he2016deep,tan2019efficientnet}. Traditional detectors are not designed for oriented objects. To make them fit the oriented object detection, we manually add an extra dimension on anchors or regression to predict the angle of the object's orientation. We denote the adaptation as `OBB' (oriented bounding box) by the end of detector's names in Table~\ref{tab:det}. To highlight the benefit from temporal modeling, we add the temporal input to baselines where 'T.' indicates the input with two successive frames and 'Ours-w/o TRL' is architecturally equivalence to the CenterPoint model with temporal input. For multiple object tracking, we include CenterTrack~\cite{zhou2020tracking} on oriented objects that use the same tracking heuristics with us for comparison.

\vspace{-5mm}
\paragraph{Implementation} We follow~\cite{sheeny2020radiate} and exclude pedestrians and groups of pedestrians from detection and tracking targets since only very few reflections are observed in these two kinds of objects. We also do not distinguish the object categories like~\cite{sheeny2020radiate} because there is no significant difference between vehicle categories presented by radar signals (\eg, truck and bus). Regarding the computation, operations related to oriented rectangles like the calculation of the overlapping of oriented bounding boxes are conducted in CPU using DOTA benchmark toolkit~\cite{xia2018dota}, while the rest part on deep neural networks is running on a single RTX 3090. For all numerical results in Table~\ref{tab:det}, we apply a center crop with size 256$\times$256 upon input images and exclude the targets outside this scope. This helps us to conduct comprehensive evaluations using our computational resource and numbers are averaged over 10 random seeds. For results in Table~\ref{tab:det_full} and ~\ref{tab:track}, we keep the original resolution with size 1152$\times$1152 to make a fair comparison to the results from~\cite{sheeny2020radiate}. We set the gap of frames between two successive frames to 3 for detection and 1 for tracking, the position dimension $D_p$ to 64, the number of temporal relational layers to 2, the batch size to 64 for cropped images with a gradient accumulation to every 2 steps, the learning rate to 5e-4 and weight decay to 1e-2 for Adam optimizer with five training epochs.

We adopt mean Average Precision (mAP) with Intersection over Union (IoU) at $0.3$, $0.5$, and $0.7$ for the evaluation of oriented object detection. For multiple object tracking, we adopt the series of MOT metrics~\cite{milan2016mot16} including MOTA, MOTP, IDSW, Frag., MT and PT, but defer the descriptions to Appendix B due to the page limitation.

\begin{table*}
  \caption{Experimental results of multiple object tracking on \textit{Radiate} dataset. TRL is the abbreviation of `temporal relational layer.'}\vspace{-3mm}
  \label{tab:track}
  \centering
  \setlength\tabcolsep{10pt}
  {
  \begin{tabular}{lcccccc}
    \toprule
    
    
    
    
    
    split: train good weather & MOTA$\uparrow$ & MOTP$\uparrow$ & IDSW$\downarrow$ & Frag.$\downarrow$ & MT$\uparrow$ & PT$\uparrow$ \\
    
    \midrule
    
    CenterTrack-ResNet18 & 0.1301 & 0.7026 & 873 & 920 & 269 & 254 \\
    CenterTrack-ResNet34 & 0.1455 & 0.7005 & 802 & 831 & 282 & 279 \\
    Ours-ResNet-18-w/o TRL & 0.3293 & 0.7135 & 513 & 593 & 151 & 324 \\
    Ours-ResNet-18-w. TRL & 0.3359 & \textbf{0.7349} & \textbf{349} & \textbf{498} & \textbf{145} & 330 \\
    Ours-ResNet-34-w/o TRL & 0.3569 & 0.7080 & 557 & 640 & 179 & 362 \\
    Ours-ResNet-34-w. TRL & \textbf{0.3791} & 0.7188 & 474 & 527 & 219 & \textbf{332} \\
    
    \bottomrule
  \end{tabular}\vspace{-4mm}
  }
\end{table*}

\begin{figure*}[ht]
    \centering
    \includegraphics[width=1.6\columnwidth]{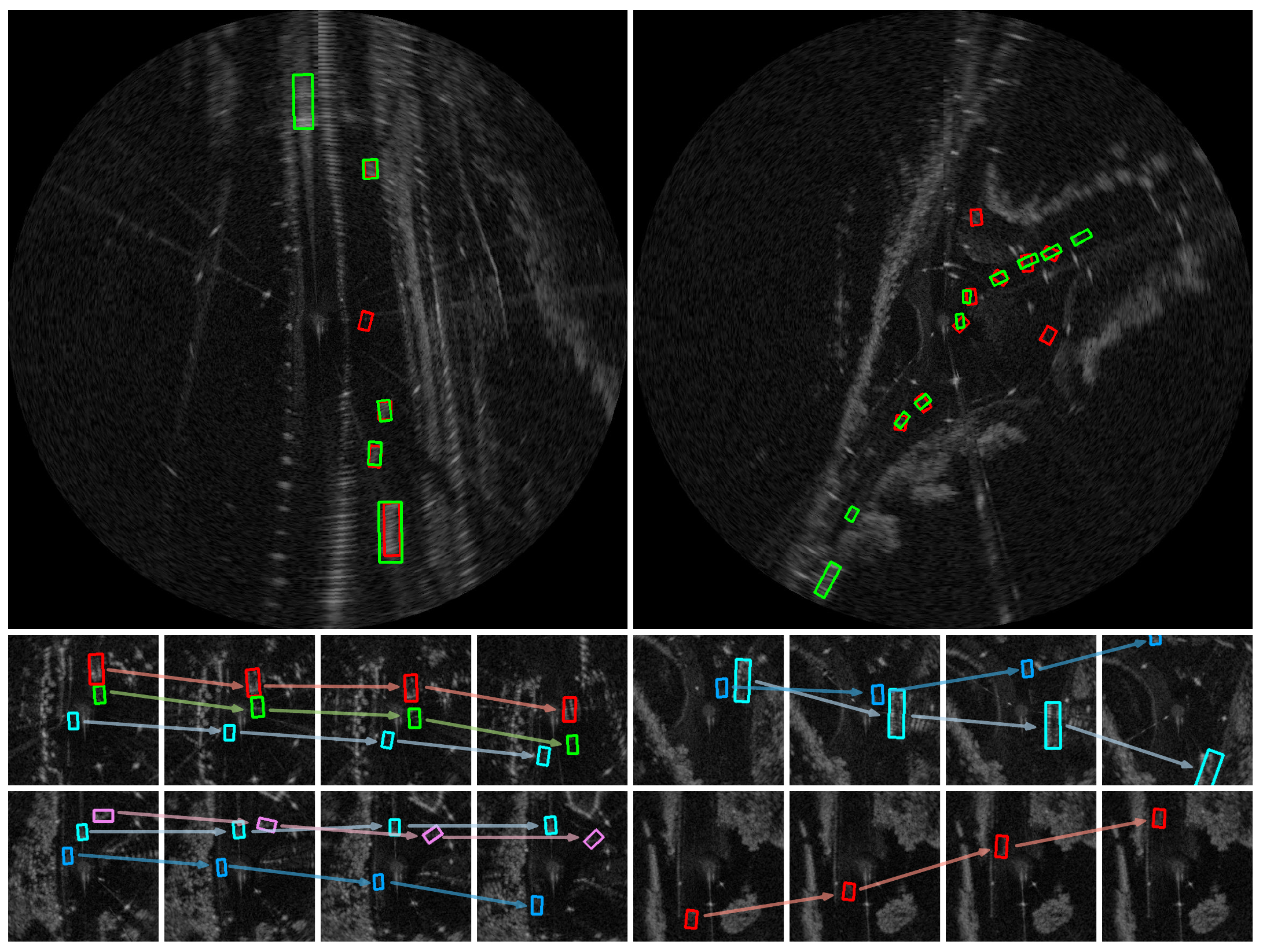}\vspace{-3mm}
    \caption{Visualizations on radar perception on \textit{Radiate} dataset. The upper two figures show the object detection while the lower four sets of successive visualizations show multiple object tracking. In detection, green bounding boxes are ground-truth annotations, while red are model predictions. In multiple object tracking, bounding boxes are model predictions, colors indicate the object IDs, and plotted arrows show the moving of objects. Regarding the figure source, the left detection figure is from night-1-4, while the right one is from rain-4-0. From left to right and top to bottom, the tracking sequences are from city-7-0, rain-4-0, fog-6-0, and junction-1-10.}\vspace{-4mm}
    \label{fig:vis}
\end{figure*}


\subsection{Result and Analysis}

\paragraph{Detection} We report detection results in Table~\ref{tab:det} and ~\ref{tab:det_full}. Our method consistently achieves better results on both two training splits among different levels of IoU thresholds. Besides, the margin between the performance with or without tempporal relational layers further confirms the contribution from modeling the temporal object consistence in successive frames. Regarding the two training splits, intuitively, adding more weather conditions into training could enhance the robustness of detection and tracking, since the testing set contains various weather. However, for radar, there is no significant difference in the presentation of data among diverse weather. The margin between two training splits mainly comes from the margin of the number of training samples. Regarding the difference in image size, there is a slight performance drop when involving a larger scope for detection. The drop comes from the cross-range resolution, where further objects might suffer from a heavier blurriness.

\vspace{-4mm}
\paragraph{Tracking} We report results on multiple object tracking in Table~\ref{tab:track}, where our methods achieve better performance comparing to baseline. For the baseline method, CenterTrack also considers the temporal information by adding the heatmap of the previous frame and the previous image into input during the inference stage. They use the ground-truth heatmap for training and the predicted heatmap for inference. This kind of learning can work well for RGB video tracking since the detection is mostly accurate. However, the detection on radar cannot achieve such accuracy so far, and therefore breaking the alignment of the heatmap in training and inference. The tracking performance with or without temporal relational layers highlights the effectiveness of modeling temporal object-level relations.

\vspace{-4mm}
\paragraph{Visualization}

We present visualization results in Fig.~\ref{fig:vis} on both object detection and multiple object tracking, and more visualizations are attached in Appendix C. We observe many predictions hit the annotations with a slight shift. Except the correct predictions, it is noticeable that our model brings some false positive predictions. However, when looking into these false positives, with a high probability, they will be a cluster of reflections inside the box that can be viewed as a ghost object. This may be the main reason for creating these false positives. Meanwhile, our model miss some objects in the outer space. The reflections of missed objects are drowning in the reflections of static surroundings due to the low angular resolution. How to enhance the detection on ghost objects and blurriness would be an interesting problem.

We add an experiment in Appendix D to analyze the best amount of selective features in temporal relational layers. The empirical results guide the heuristic setting of $K$.


\section{Related Work}\label{sec:relate}

\paragraph{Radar Perception in Autonomous Driving} There is an increasing attention on the adoption of radar in autonomous driving. We review some recent work from both algorithmic and radar resource perspectives. The work~\cite{major2019vehicle} proposes a deep-learning approach for automotive radar object detection using range-azimuth-doppler measurement. \cite{qian2021robust} focus on sensor fusion and propose a method to incorporate synchronous radar and Lidar signals for object detection. ~\cite{yang2020radarnet,lim2019radar} also exploit the multi-modal sensing fusion in autonomous driving. Besides deep learning, Bayesian learning has also been used for extended object tracking using radar~\cite{yao2021extended,xia2021learning}. Our work only leverages radar signals but enhances the recognition with the temporal consistency on objects, which has not been explored by previous works. We defer a short review of current radar dataset in Appendix E.

\vspace{-4mm}
\paragraph{Detection with Temporality} Consecutive video frames could provide spatial-temporal cues for object recognition. \cite{wu2019long} leverage a feature bank that extends the time horizon for spatial-temporal action localization. \cite{shvets2019leveraging} and \cite{beery2020context} insert the object-level association from short or long temporal dependency into Faster-RCNN~\cite{ren2015faster} to capture the spatial-temporal information in object detection. Other techniques such as video pixel flow or 3D convolutions~\cite{zhu2017deep,zhu2017flow,xie2018rethinking} are applied for visually rich video sequences but too heavy and not efficient for radar images. Our work shares the same philosophy that using spatial-temporal object-level correlation along the time horizon. However, all studies mentioned above are focusing on RGB video data but not design for oriented objects. The object's size and scale may not be consistent if an object is approaching or leaving the scope of the camera. Differently, we put our emphasis on radar data in autonomous driving, where the bird-eye-view point cloud-based images provide significant object property comparing to RBG video data. We design an anchor-free one-stage detector with temporality, which is efficient and does not have to tackle the pre-defined anchor parameters. The center-based detector is suitable for the bird-eye-view presentation since there is no object overlap from this view, hence the central feature is fully exposed to represent an object. Moreover, we do not explore the long-range dependency but restrict the consistency in only one successive frame, since vehicles can move out of the scope if the timescale is too long and consequently no more temporal relation is available.

\vspace{-4mm}
\paragraph{Multiple Object Tracking} A well-established paradigm for visual multiple object tracking~\cite{milan2016mot16} is tracking-by-detection~\cite{jiang2019graph,schulter2017deep,tang2017multiple}. The detected object bounding boxes are provided by an external detector, then data association techniques based on object appearance or motion are applied to detection to associate identical objects among candidates in multiple consecutive frames. Recent developments in multiple object tracking convert detectors into tracking algorithms to jointly detect and track objects~\cite{zhou2020tracking,feichtenhofer2017detect,yin2021center}. We follow the simple tracking rule that is purely based on the cost of euclidean distance~\cite{zhou2020tracking,yin2021center} to extend our framework to multiple object tracking. Differently, ~\cite{zhou2020tracking,yin2021center} only stack frames at multiple time steps as input, while our networks explicitly consider the object-level consistency.


\section{Conclusion}

We studied the object recognition problem using radar in autonomous driving. We facilitated the radar perception with temporality from video frames based on the assumption that the same object within successive frames should be consistent and share almost the same attributes. We designed a framework inserted with temporal relational layers to explicitly model the object-level consistency. We showed the effectiveness of our method by experiments in object detection and multiple object tracking.

\vspace{-4mm}
\paragraph{Acknowledgement} The authors would like to thank Petros T. Boufounos, Toshiaki Koike-Akino, Hassan Mansour, and Philip V. Orlik for their helpful discussion.

\clearpage
{\small
\bibliographystyle{ieee_fullname}
\bibliography{egbib}
}


\clearpage
\appendix

\section{Temproal Feature Extraction}

\begin{figure*}[ht]
    \centering
    \includegraphics[width=1.9\columnwidth]{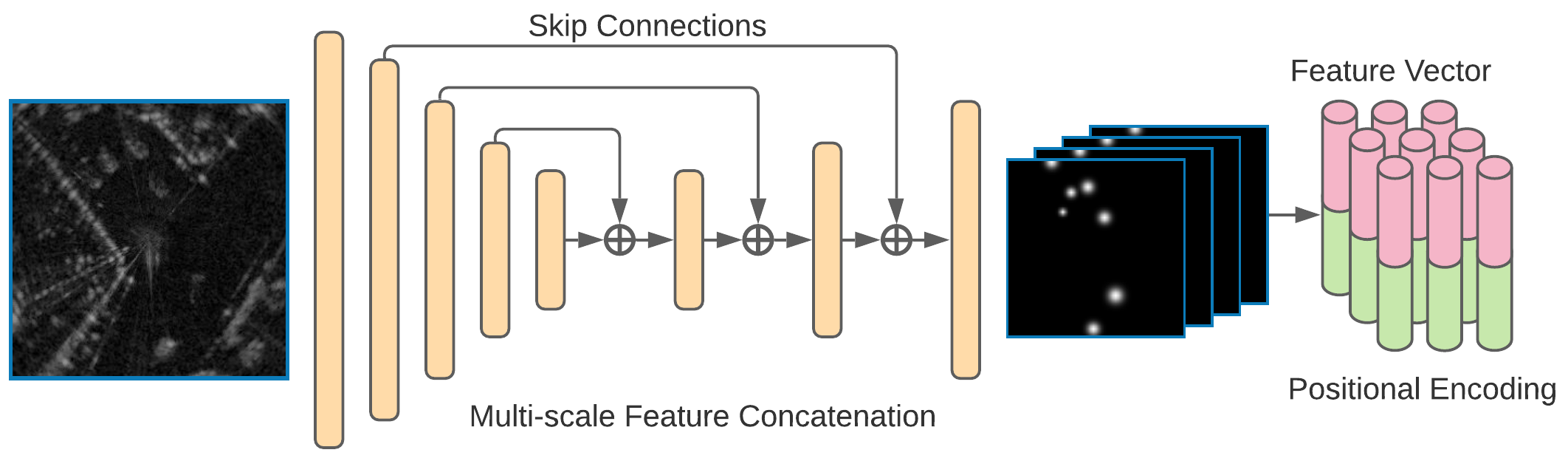}
    \caption{The backbone networks are inserted with several skip connections to collect features at different scales for predictions. Features selected for temporal relations modeling are attached with positional encoding to reveal the locality of objects.}
    \label{fig:extraction}
\end{figure*}

We add Fig.~\ref{fig:extraction} to illustrate the skip connections in the backbone neural networks. Skip connections within CNNs are designed to jointly involve high-level semantics and low-level finer details in output feature representation. Specially, we add three skip connections in ResNet and gradually up-sample the features from a deeper layer. The final feature representations are down-sampled with a ratio of $4$ compared to the original inputs in this U-Net structure.


\section{Multiple Object Tracking: Evaluation and the Decoding Algorithm}

We adopt the series of MOT metrics~\cite{milan2016mot16} for evaluation. We pick several key metrics in experiments: MOTA (Multiple Object Tracking Accuracy), MOTP (Multiple Object Tracking Precision), ID switch (IDSW), track fragmentations (Frag.), mostly tracked (MT), and partially tracked (PT). The MOTA score is calculated by
\begin{equation*}
    \text{MOTA} = 1 - \frac{\sum_t (\text{FN}_t + \text{FP}_t + \text{IDSW}_t)}{\sum_t \text{GT}_t},
\end{equation*}
where $t$ is the frame index, GT is the number of ground-truth objects, FN and FP refer to false negative and false positive detection. The value of MOTA is in the range $(-\infty, 100]$. It can be deemed as the combination of detection and tracking performance, and is widely used as the main metric for accessing multiple object tracking quality. MOTP is the average IoU value on all ground-truth bounding boxes and its assigned prediction. It describes the localized precision. The rest of these metrics all reflect the quality of predicted tracklets. For detailed definitions and calculations of MOT metrics, please refer to~\cite{milan2016mot16}.

We attach a decoding algorithm for multiple object tracking. The tracking algorithm mainly follows~\cite{zhou2020tracking} which associates objects from successive frames purely based on the cost of Euclidean distance. The position of an object in the previous frame is complemented with a predictive positional tracking offset $\hat{\textbf{d}}$ to infer its potential position in the next frame. Then, objects in previous and current frames are associated and propagate the object's ID in a bipartite graph with a greedy algorithm based on the distance between their center 2D positions. Empirically, we do not further extend a tracklet if it cannot find a matched candidate.

\begin{algorithm}[ht]
\caption{Multiple Object Tracking Decoding}\label{alg:track}
\begin{algorithmic}[1]
\Require $T^{t-1} = \{(\textbf{c}, \text{id})_j^{t-1}\}_{j=1}^M$: tracked objects in the previous frame $t-1$; $\hat{B}^t = \{(\hat{\textbf{c}}, v, \hat{\textbf{d}})_i^t\}_{i=1}^N$ heatmap predictions of object centers $\hat{\textbf{c}}$, confidence $v$, and tracking offsets $\hat{\textbf{d}}$. $\hat{B}^t$ are sorted in a descending order according to $v$. Distance threshold $k$. Birth threshold $b$.
\State $S\gets \emptyset$, $T^t \gets \emptyset$
\State $W\gets \text{Cost}(\hat{B}^t, T^{t-1})$ \Comment{$W_{ij} = || \hat{\textbf{c}}_i^t - \hat{\textbf{d}}_i^t, \hat{\textbf{c}}_j^{t-1} ||_2$}

\For{$i\gets 1, N$}
    \State $j\gets \argmin_{j\notin S} W_{ij}$
    \If{$w_{ij}\leq k$}
        \State $T^t\gets T^t\cup (\hat{\textbf{c}}_i^t, \text{id}_j^{t-1})$ \Comment{Propagate matched id}
        \State $S\gets S\cup \{j\}$ \Comment{Mark candidate $j$ as tracked}
    \ElsIf{$v_i\geq b$}
        \State $T^t\gets T^t\cup (\hat{\textbf{c}}_i^t, \text{New id})$ \Comment{Create a new track}
    \EndIf
\EndFor

\State \textbf{return} $T^t$
\end{algorithmic}
\end{algorithm}


\section{Ablation Study}

\begin{figure}
    \centering
    \includegraphics[width=\columnwidth]{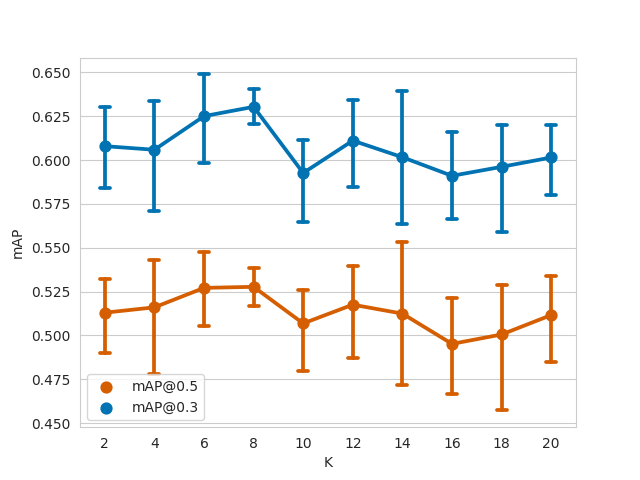}
    \caption{Detection performance with varying $K$ value.}
    \label{fig:k}
\end{figure}

We add an experiment on split train good weather in Fig.~\ref{fig:k} to analyze the change of the number of the selective feature vectors for temporal relational layers, where we vary the value $K$ from 2 to 20. The detection performance consistently improved before $K$ reached 8, but drop when continually increase the value of $K$. The scenario indicates involving redundant objects in relation modeling could slightly corrupt the temporal relation learning. The value of $K$ should be selected based on the average number of objects per frame but not including excessive noise. We empirically set $K$ to 8 in our experiments.

\section{Additional Visualization Result}

\begin{figure*}
    \centering
    \includegraphics[width=1.9\columnwidth]{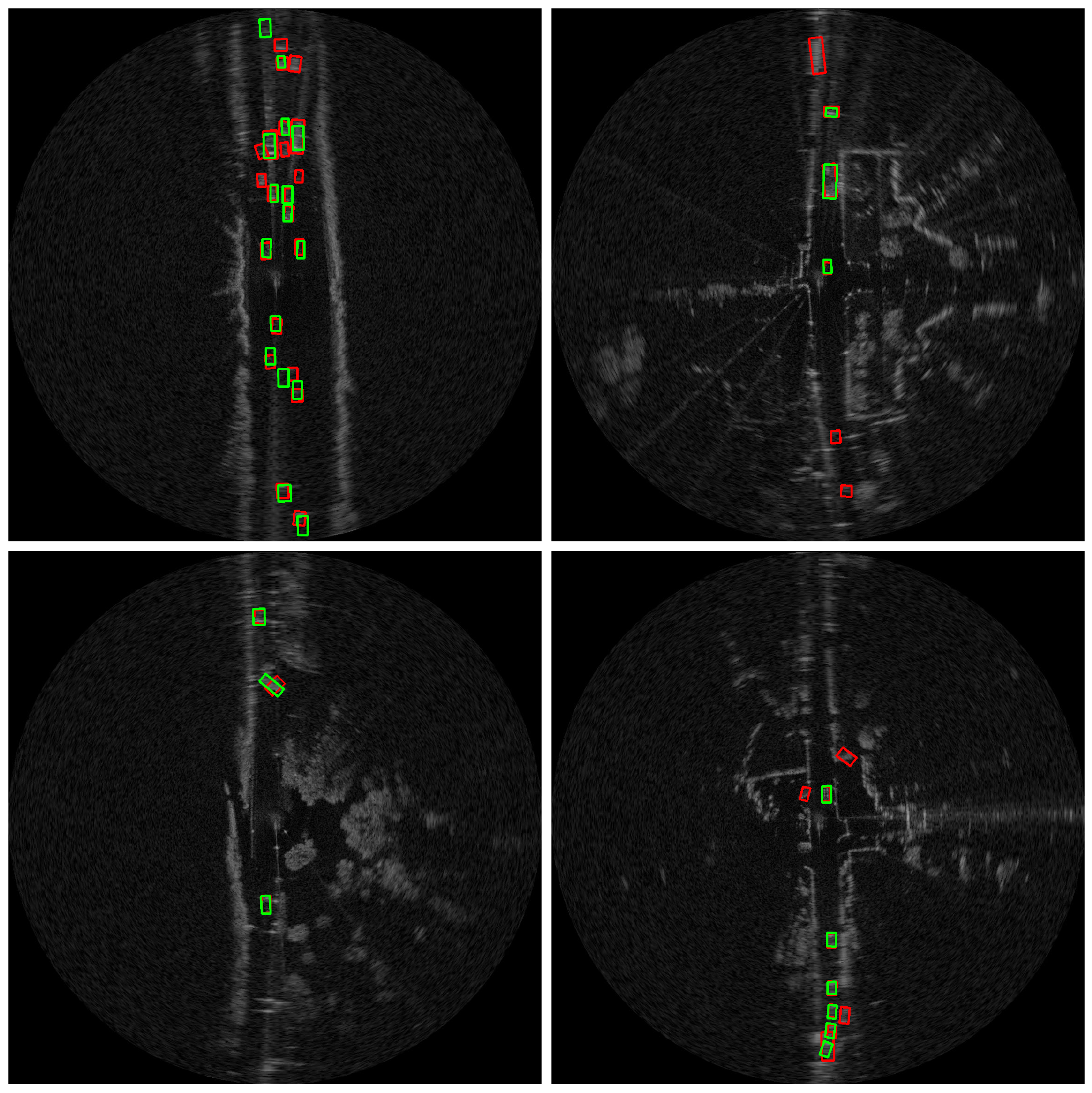}
    \caption{Visualizations on object detection. From left to right and top to bottom, the figures are from: motorway-2-1, tiny foggy, junction-1-10, and fog-6-0. Green bounding boxes are ground-truth annotations, while red are model predictions from `Ours-ResNet18-w. TRL.'}
    \label{fig:add_vis}
\end{figure*}

We present additional visualization results in Fig.~\ref{fig:add_vis} on object detection. In the detection, green bounding boxes are ground-truth annotations, while red are predictions. The same observations are confirmed in the additional visualizations. False positive predictions are mainly due to the `ghost' objects in radar signals, and the rest are localized in the surroundings or outer space where the angular resolution is low.

\section{A Short Review of Radar Dataset}

Besides the algorithmic design, many radar datasets are emerging which are crucial for machine learning research. Among these datasets, radar data are currently presented in various data formats, \ie radio frequency heatmap, radar reflection image, or point cloud. \textit{RadarScenes} dataset~\cite{schumann2021radarscenes} provide abundant point-wise annotations with doppler for automotive radar. However, there is no bounding box annotation for objects. \textit{Carrada} dataset~\cite{ouaknine2021carrada} records the range-angle and range-Doppler heatmap. Their data are mainly recorded in experimental sites like parking lots but not in real driving environment. \textit{CRUW} dataset~\cite{wang2021rethinking} offers radar's radio frequency images with camera-projected annotations. \textit{nuScenes}~\cite{caesar2020nuscenes} contains multi-modal data including Lidar, camera, and radar. However, radar data in \textit{nuScenes} only afford sparse point cloud, while the Lidar and camera data are the main advantage of this dataset. \textit{MulRan}~\cite{kim2020mulran} and \textit{Oxford}~\cite{barnes2020oxford} datasets present high-resolution radar images for urban driving scenarios but without object-level annotation. In our paper, we conduct detection and tracking experiments on point cloud-based radar images in adverse weather from \textit{Radiate} dataset~\cite{sheeny2020radiate}, and every significant object has bounding box and tracking ID annotations for training.

\end{document}